\documentclass[runningheads]{llncs}
\usepackage[T1]{fontenc}
\usepackage{graphicx}
\usepackage{amsmath} 
\usepackage{booktabs}
\usepackage{multirow}
\usepackage{cite}
\usepackage[hidelinks]{hyperref}

\begin{document}
%

\title{Few-shot Writer Adaptation via \texorpdfstring{\\}{ }Multimodal In-Context Learning}

\titlerunning{Few-shot Writer Adaptation via MICL}

\author{Tom Simon\inst{1}\and Stéphane Nicolas\inst{1} \and
Pierrick Tranouez\inst{1} \and Clément Chatelain\inst{2} \and
Thierry Paquet \inst{1}}
\authorrunning{T. SIMON et al.}

\institute{LITIS EA4108, University of Rouen Normandy, France
\email{\{tom.simon,stephane.nicolas, pierrick.tranouez, thierry.paquet\}@univ-rouen.fr}\\ 
\and LITIS EA4108, INSA of Rouen Normandy, France
\email{clement.chatelain@insa-rouen.fr}
}
\date{January 2026}

\maketitle

\vspace{-4mm}
\begin{abstract}
While state-of-the-art Handwritten Text Recognition (HTR) models perform well on standard benchmarks, they frequently struggle with writers exhibiting highly specific styles that are underrepresented in the training data. To handle unseen and atypical writers, writer adaptation techniques personalize HTR models to individual handwriting styles. Leading writer adaptation methods require either offline fine-tuning or parameter updates at inference time, both involving gradient computation and backpropagation, which increase computational costs and demand careful hyperparameter tuning. In this work, we propose a novel \textbf{context-driven HTR framework}\footnote{The source code will be made publicly available upon publication.} inspired by multimodal in-context learning, \textbf{enabling inference-time writer adaptation} using only a few examples from the target writer \textbf{without any parameter updates}. We further demonstrate the impact of context length, design a compact 8M-parameter CNN-Transformer that enables few-shot in-context adaptation, and show that combining context-driven and standard OCR training strategies leads to complementary improvements. Experiments on IAM and RIMES validate our approach with Character Error Rates of 3.92\% and 2.34\%, respectively, surpassing all writer-independent HTR models without requiring any parameter updates at inference time.
\end{abstract}
\vspace{-6mm}

\vspace{-2mm}
\section{Introduction}
\begin{figure}[h]
  \centering
  \resizebox{\textwidth}{!}{\includegraphics{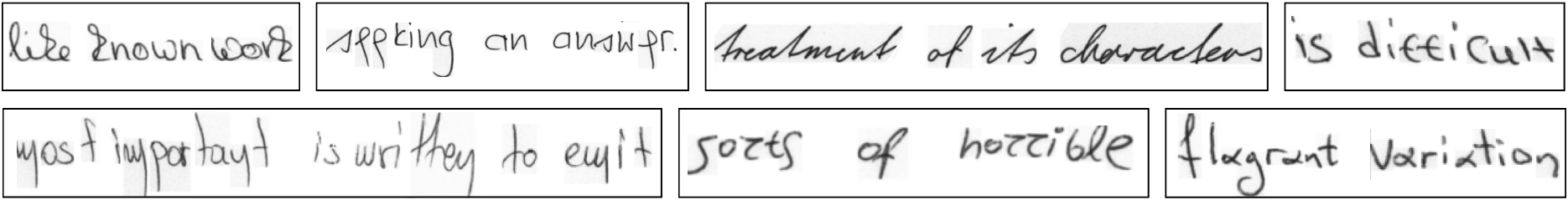}}
\vspace{-6mm}
\caption{Qualitative examples of inter-writer variation in the IAM dataset, highlighting writer-specific characteristics in letter formation.}
\label{fig:inter_writer_variation}
\end{figure}

\vspace{-1mm}
Over the past decade, Handwritten Text Recognition (HTR) has made remarkable progress, with state-of-the-art systems achieving strong performance on standard benchmarks~\cite{iam,rimes}. Despite these advances, modern HTR models often struggle with rare or highly distinctive handwriting styles that are underrepresented in the training data. Some writers develop distinctive ways of shaping certain letters, sometimes in a completely unique manner that is absent from the training data, creating significant ambiguity and performance degradation in HTR systems (see Fig.~\ref{fig:inter_writer_variation}). The sensitivity of HTR models to previously unseen writers represents a significant limitation in real-world scenarios, where handwriting styles can vary widely, highlighting the necessity of \textbf{writer adaptation} strategies. 

Formally, writer adaptation aims to specialize a writer-independent model — originally trained on large-scale, multi-writer datasets — by adjusting it to reflect the morphological and stylistic characteristics of a particular individual's script. To address this challenge, several strategies have emerged, ranging from supervised fine-tuning to dynamic inference-time updates. A common approach consists of offline fine-tuning, assuming prior access to labeled data from the target writer \cite{soullard2019,pippi2023choose}. While effective, this specialization requires a sufficient amount of labeled data~\cite{read} and introduces the risk of overfitting to a particular writing style, thereby reducing the model’s ability to generalize to unseen styles. To address these limitations, more recent works have explored \textbf{few-shot and test-time adaptation strategies}, where adaptation is performed at inference time using a small support set (typically fewer than ten samples) \cite{metahtr,wang2022fast}. These approaches leverage either explicit style representations \cite{wang2022fast,kohut2023towards} or reconstruction-based objectives, including Masked Autoencoder frameworks \cite{docttt,metawriter}. However, the best-performing writer-adaptation approaches typically rely on gradient-based parameter updates at inference time. This requirement introduces significant architectural complexity and memory overhead, as it necessitates the maintenance of optimization states during deployment. 

\begin{figure}[h]
  \centering
  \resizebox{\textwidth}{!}{\includegraphics{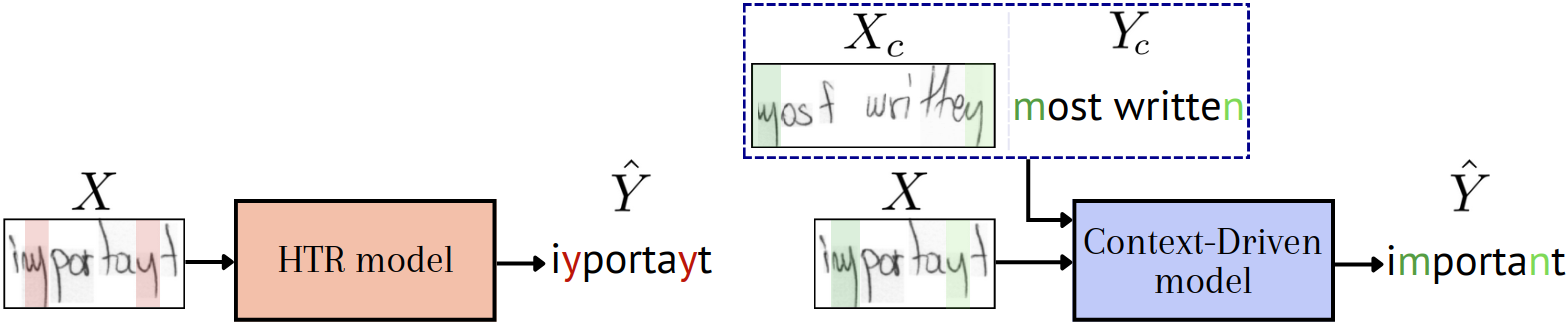}}
\vspace{-6mm}
\caption{By integrating context lines \(X_c\) and their corresponding labels \(Y_c\) from the same author as the query \(X\), our context-driven model effectively captures the writer's stylistic characteristics. Leveraging these few in-context examples significantly reduces errors caused by writer-specific variations.}
  \label{fig:high_level}
\end{figure}

In this work, we propose a novel context-driven HTR framework inspired by recent advances in multimodal in-context learning \cite{flamingo,e2str,rosetta}. Our approach achieves inference-time writer adaptation using only a few in-context examples from the target writer, \textbf{eliminates the need for parameter updates, and outperforms strong HTR baselines} (see Fig.\ref{fig:high_level}). 

\vspace{-5mm}

\vspace{-10mm}
\newpage
While previous models leveraging multimodal in-context learning techniques were based on architectures with more than 100 million parameters~\cite{e2str,rosetta}, we design a \textbf{compact 8 million parameters CNN-Transformer model} that enables \textbf{few-shot in-context adaptation} and remains robust and effective even with a lightweight architecture. Furthermore, we provide an extensive empirical analysis of the impact of context length, showing consistent performance gains as the context increases from 1 to 9 lines. We demonstrate that fusing context-driven and standard OCR prediction enhances recognition accuracy and confirm its effectiveness on standard benchmarks, achieving a Character Error Rate (CER) of 3.92\% on IAM and 2.34\% on RIMES, surpassing existing state-of-the-art HTR models.

\vspace{-3mm}
\section{Related Work}
\vspace{-1mm}
This section first reviews the evolution of HTR systems and the landscape of writer-adaptation techniques. We then discuss recent breakthroughs in multimodal in-context learning, which we leverage to build our context-driven HTR framework for parameter-free writer adaptation.
\vspace{-2mm}
\subsection{Handwritten Text Recognition}
Handwritten Text Recognition (HTR) converts handwritten text into machine-readable sequences using deep neural networks. It introduces several challenges: high morphological variability across and within writers, cursive or connected writing with unclear character boundaries, and stylistic inconsistencies such as irregular spacing and letter slant. HTR architectures have evolved from CNN-RNN pipelines 
\cite{crnn,puigcerver2017,peroocr,bestpractices} to hybrid CNN-Transformer \cite{dan,daniel} and fully Transformer-based models \cite{donut,dessurt,trocr,dtrocr}. While the latter achieve state-of-the-art performance through large-scale pretraining, they require hundreds of millions of parameters and substantial computational resources. Furthermore, many HTR systems incorporate linguistic priors that may bias predictions toward plausible words rather than visually grounded evidence \cite{language_model_issues}, which is particularly problematic given the 
high stylistic variability of handwritten text. Despite these advances, HTR models operate with fixed weights and lack mechanisms to adapt to writer-specific variations, failing when confronted with highly distinctive styles that are underrepresented or absent in the training data — motivating the need for dedicated writer adaptation strategies.

\vspace{-2mm}
\subsection{Writer Adaptation}
\label{sec:writer_adaptation}
\vspace{-1mm}
Three main strategies exist for adapting to diverse writer's styles: (1) improving training-time generalization, (2) offline fine-tuning for writer specialization, and (3) inference-time writer adaptation.

\paragraph{Training-Time Generalization.}
A first line of work improves robustness to unseen writing styles through training. Data augmentation has become a standard component of HTR training pipelines, including geometric transformations, elastic distortions, and noise injection~\cite{jaderberg2014,wigington2017}. Beyond classical augmentation, synthetic data generation offers a complementary direction to expand style diversity: font-based rendering approaches~\cite{dan,vlt,daniel} generate large-scale training sets at low cost, while Handwritten Text Generation (HTG) models~\cite{wordstylist,vatr} produce more realistic samples by learning to mimic real writing styles.  However, both directions face limitations: font-based synthetic data suffers from a persistent domain gap~\cite{kang2020unsupervised}, while HTG-generated data can degrade HTR performance without careful filtering. Only a fraction of samples consistently improves recognition~\cite{wordstylist}.

\paragraph{Offline Fine-Tuning Adaptation.} While training-time strategies enhance global robustness, offline fine-tuning assumes prior access to labeled data from the target writer, enabling explicit model specialization before deployment.  Soullard et al.~\cite{soullard2019} demonstrate that fine-tuning both the optical and language models with as few as 1 to 16 annotated pages from the target writer yields significant CER improvements. Pippi et al.~\cite{pippi2023choose} complement this by providing practical guidelines on how to select the most appropriate pretrained model before fine-tuning, showing that the choice of source domain significantly impacts single-writer adaptation performance.  Fine-tuning remains the most widely adopted and effective strategy for writer-specific adaptation~\cite{soullard2019,pippi2023choose,kohut2023fine}, as it directly specializes model parameters towards the target writing style. However, this specialization comes at the risk of over-fitting to a specific style, reducing the model's ability to handle style variability at inference time.

\paragraph{Inference-Time Adaptation.}
In contrast to offline methods, inference-time adaptation methods adapt to a specific writer on the fly, leveraging a few examples from the target writer at test time. Parameter-free approaches condition the model on writer-specific representations extracted from unlabeled contextual examples~\cite{wang2022fast,kohut2023towards}, however, their effectiveness on standard benchmarks remains limited. The top-performing writer adaptation methods in the literature  update a subset of parameters through gradient-based optimization at inference time, leveraging self-supervised reconstruction objectives based on Masked Autoencoders (MAE)~\cite{metahtr,docttt,metawriter}. However, the connection between image reconstruction and writer-specific recognition remains implicit and difficult to interpret: \textbf{reconstructing a noised or masked image does not necessarily resolve the visual ambiguities inherent in a writer's style} — a model can learn to accurately reconstruct handwritten text without ever learning to disambiguate characters that are systematically confused due to writer-specific morphological variations. In other words, reducing reconstruction error and reducing character-level ambiguity are fundamentally different objectives, and there is no guaranty that optimizing the former directly addresses the latter. Moreover, in practice, the effective use of MAE in test-time adaptation is non-trivial: when the reconstruction loss is not properly balanced with the transcription objective, reconstruction gradients can conflict with the primary learning signal and degrade recognition performance~\cite{docttt,metawriter}. The masking procedure itself introduces an additional sensitivity, as varying the masking ratio or strategy directly controls the difficulty of the reconstruction task, which in turn affects the magnitude of the reconstruction loss and its gradient signal, creating a cascading effect where multiple interdependent hyperparameters — masking ratio, loss weighting, and learning rate — must be carefully co-tuned to avoid either trivial reconstruction that provides no useful adaptation signal or overly aggressive updates that destabilize the transcription objective.
In this work, we explore an alternative approach inspired by Multimodal In-Context Learning (MICL). 
\vspace{-2mm}
\subsection{Multimodal In-Context Learning}
\vspace{-1mm}
In-Context Learning (ICL), first introduced in large language models, enables models to perform new tasks by leveraging a set of input-output examples (\(X_c, Y_c\)) provided in the input context, without requiring parameter updates~\cite{gpt3,icl_explanation}. Multimodal In-Context Learning (MICL) extends this paradigm to vision-language models by conditioning predictions on joint visual and textual in-context examples~\cite{flamingo,openflamingo,idefics,mmicl,mm1,blip3}. This capability allows models to adapt dynamically to new tasks and domains at inference time. MICL has shown strong performance in various multimodal tasks, including image classification, image captioning, visual question answering, and video question answering, highlighting it as a flexible and general paradigm. Despite these advances, applications of MICL to text recognition remain limited~\cite{e2str}. Furthermore, recent studies have shown that MICL models often exhibit an imbalance in modality utilization, relying heavily on textual patterns while under-exploiting visual information~\cite{truemicl,whatmakes}. This limitation is particularly critical in handwritten text recognition, where accurate transcription relies on fine-grained visual features. As a result, \textbf{the proper exploitation of both modalities remains an open challenge for deploying MICL methods in real-world HTR scenarios}.

Rosetta~\cite{rosetta} recently proposed a context-driven training strategy that enhances visual grounding in MICL for text and symbol recognition. The core idea is to condition the prediction \(\hat{Y}\) of a query text image \(X\) on its current context, leveraging a dedicated text encoding. This is achieved through a Context-Aware Tokenizer (CAT), which converts each character in the context transcription \(Y_c\) into a relative position token within a sequence \(T\). Unlike standard HTR models, which are trained to predict character classes directly, Rosetta trains the model to indicate, for each character in \(X\), whether it appears in the current context \(X_c\) and, if so, to associate it with the corresponding class in \(T_c\). If a character is absent from the context, the model is trained to predict a special out-of-context token. Consequently, the model cannot rely on fixed pattern-class memorization as in conventional HTR; instead, \textbf{it is required to dynamically leverage the provided context (\(X_c, Y_c\)) in order to correctly predict the transcription \(Y\)}. By conditioning character recognition on their visual and textual context, the model has several key advantages: (1)it naturally grounds its predictions in relation to the visual modality of the context examples, mitigating the textual bias observed in standard MICL approaches; (2)it generalizes across alphabets and writing systems.
While promising, Rosetta's prior evaluation was \textbf{limited to clean synthetic printed data} with minimal geometric variations, where character shapes are stable, with well-defined boundaries and separations between neighboring characters, highlighting a significant gap with handwritten text~\cite{kang2020unsupervised}. 
Furthermore, the context length was highly restricted, limited to a single line, and the model employed in their experiments contained 240 million parameters, substantially larger than contemporary specialized HTR models~\cite{span,decoupledan,kang2022pay,van,dan,vlt}.

In this work, we adapt and extend the context-driven recognition paradigm of Rosetta to leverage MICL for writer adaptation in real-world handwritten text recognition — a significantly more challenging setting than those previously addressed by Rosetta, involving cursive writing, ambiguous character boundaries, and high intra- and inter-writer variability. Unlike MAE-based methods, this paradigm offers a more interpretable adaptation objective directly tied to visual pattern matching rather than implicit reconstruction. We propose a compact 8-million-parameter CNN-Transformer architecture that enables effective \textbf{few-shot writer adaptation 
through multimodal in-context examples, without any parameter updates}.

\vspace{-2mm}
\section{Context-Driven Framework for HTR}
\vspace{-1mm}
We propose a context-driven architecture that adapts to a specific writer's handwriting style by leveraging a context composed of a small set of line images \(X_c\) and their corresponding transcriptions \(Y_c\) from the same writer. By exploiting the characteristic letter shapes present in these examples, the model can resolve ambiguities in a query image \(X\) and accurately predict its transcription \(Y\). Rather than learning fixed associations between visual patterns and predefined classes like conventional OCR/HTR systems, the model is trained to effectively exploit the context (\(X_c\), \(Y_c\)), and its prediction \(\hat{Y}\) is conditioned on this context (See Fig.\ref{fig:high_level}). 
For each character region in \(X\), the model retrieves visually similar patterns in \(X_c\), identifies the corresponding symbol in \(Y_c\), and appends it to \(\hat{Y}\).  As a result, the model learns a general inference strategy that can be consistently applied to previously unseen writers at test time, allowing it to adapt dynamically through the use of in-context examples. This design enables \textbf{writer adaptation at inference time without any parameter updates}.

\subsection{Model Architecture} 

\begin{figure}[h]
  \centering
  \resizebox{\textwidth}{!}{\includegraphics{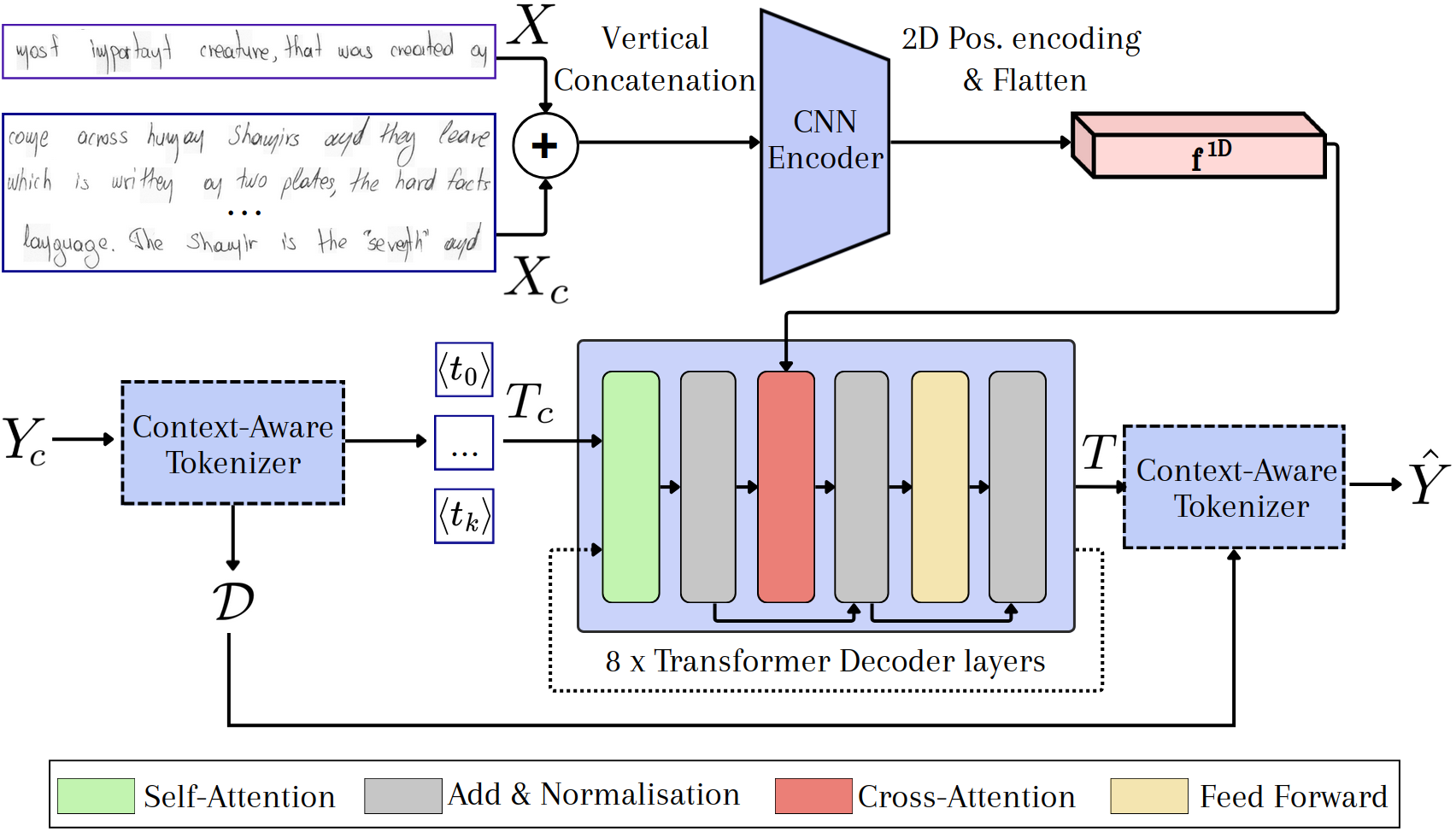}}
\caption{Illustration of our Context-driven architecture, structured around three core components: (1) a Context-Aware Tokenizer (2) a CNN encoder and (3) a Transformer decoder}
  \label{fig:architecture}
\end{figure}

Motivated by the goal of adapting OCR architectures already deployed in real-world scenarios to the in-context learning paradigm, we propose a CNN Transformer model inspired by recent HTR architectures~\cite{dan,vlt}, which we extend to incorporate in-context examples (see Fig. \ref{fig:architecture}). Our architecture leverages context lines \(X_c\) from the same writer as the query line \(X\), along with their corresponding transcriptions \(Y_c\), to enable writer-adaptation. 
It is structured around three core components: (1) a Context-Aware Tokenizer (CAT), (2) a CNN visual encoder, and (3) a Transformer decoder. Apart from the parallel processing enabled by teacher forcing in the Transformer decoder during training, the architecture operates identically during training and inference time.

\newpage
\vspace{-6mm}
The context and query images (\(X_c\),  \(X\)) are first vertically concatenated into a single composite image and processed by a CNN encoder to produce a 2D feature map. A 2D sinusoidal positional encoding is then added to preserve spatial information, and the resulting feature map is flattened into a one-dimensional feature sequence  \(f^{\mathrm{1D}}\). This design naturally supports images with varying widths and heights while avoiding distortion from resizing operations. 

The Context-Aware Tokenizer (CAT) encodes the context transcriptions \(Y_c\) into sequences of relative position tokens \(T_c\), where each unique character class is assigned a token \(\langle t_i \rangle\) according to the order of its first occurrence in the sequence. For instance, if the character \texttt{'a'} is the fourth distinct character appearing in \(Y_c\), it is encoded as \(\langle t_4 \rangle\). This bijective mapping between character labels and relative position tokens is stored in a dictionary \( \mathcal{D} \).

The Transformer decoder processes the encoded context tokens \(T_c\) via self-attention, while attending to the visual features \(f^{\mathrm{1D}}\) through cross-attention.
Through cross-attention between textual and visual representations, the decoder auto-regressively predicts the sequence of relative position tokens \(T\) for the query image. Each predicted token is then decoded by the CAT using the mapping dictionary \( \mathcal{D} \), yielding the final character sequence \(\hat{Y}\).

\vspace{-2mm}
\subsection{Training Protocol}
\label{sec:train_protocol}
\vspace{-1mm}
The model is trained to effectively exploit the context (\(X_c\), \(Y_c\)). For each character instance in the query image \(X\), if this instance appears in the context images \(X_c\), the model is trained to predict the corresponding relative position token defined by this context-dependent encoding \(T_c\), resulting in the sequence \(T\). If a character does not appear in the context, the model is explicitly trained to predict a dedicated out-of-context token \(\langle \text{ooc} \rangle\).
To facilitate progressive learning, we adopt several \textbf{curriculum learning strategies}. Initially, to reduce visual variability between context and query characters during the early stages of training, we use synthetic printed text from about 1,000 font styles applied to training-set transcriptions, without adding any external linguistic information. Real handwritten data is then gradually introduced, with the proportion of real samples increasing from 20\% to 80\% over the course of training, allowing the model to build reliable query-to-context associations under controlled visual conditions before being exposed to the full complexity of handwritten variations. In parallel, we introduce a curriculum on the context size, where the maximum number of context lines \(K\) is gradually increased, enabling faster and more stable convergence before the model is exposed to longer and more complex context sequences.
To enhance inference robustness, we introduce noise during teacher forcing by randomly replacing ground-truth tokens with alternatives from the context set $T_c$. Beyond independent token corruption, we implement a structured \textbf{group noise mechanism} where all occurrences of a given relative position token in the past sequence $T^{i-1}$ are jointly replaced by a token from $T_c$. 
This strategy prevents the model from consistently associating query characters with incorrect context tokens, thereby reducing error propagation during autoregressive decoding.

\vspace{-2mm}
\subsection{Late Fusion of Standard OCR and Context-Driven Models}
\label{sec:fusion}
\vspace{-1mm}
While a standard OCR model uses its language knowledge to handle visually ambiguous characters, the Context-Driven model relies on in-context examples to guide its predictions. Motivated by the complementary strengths of these models (see Section \ref{sec:complementary}), we adopt a fusion strategy that combines the predictions from these two models. We train two independent models built upon the same CNN–Transformer backbone architecture, but optimized under distinct training paradigms: (1) standard OCR training and (2) context-driven training. While the Context-Driven model incorporates in-context examples during training, the standard OCR model is trained exclusively on isolated handwritten text line images, without access to any additional contextual information and without using the CAT. 

\vspace{-3mm}
\begin{figure}[h]
  \centering
  \resizebox{\textwidth}{!}{\includegraphics{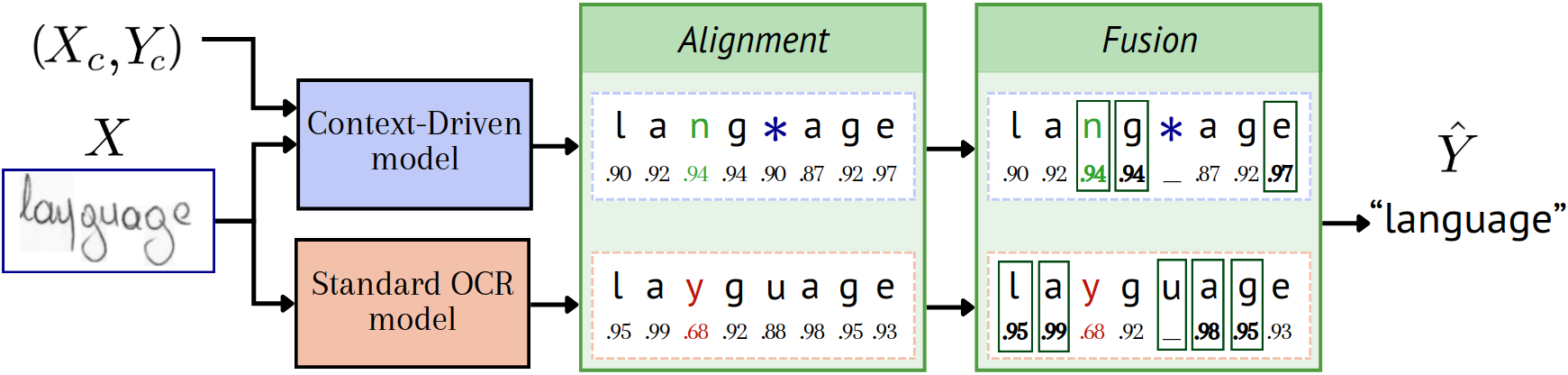}}
\caption{Confidence-based fusion of Context-Driven and Standard OCR predictions. $\langle \text{ooc} \rangle$ prediction is represented by the symbol '*'}
\label{fig:fusion}
\end{figure}
\vspace{-3mm}

As illustrated in Fig.~\ref{fig:fusion}, at inference time, both models output class predictions along with the associated confidence scores at the character level, computed by applying a softmax to the logits (raw network outputs). Both models employ greedy decoding without any external language model, ensuring that predictions rely solely on visual evidence and learned representations rather than linguistic priors. To fuse the predictions, we first align the two predicted sequences by minimizing their Levenshtein distance via dynamic programming, recovering the optimal character-level alignment. At each aligned position, we select the prediction with the highest confidence score. In cases where the Context-Driven model predicts an out-of-context token $\langle \text{ooc} \rangle$, the corresponding OCR prediction is used instead, regardless of confidence. 

\vspace{-2mm}
\section{Experiments and Results}

\subsection{Datasets and Setup}
\label{sec:setup}
We evaluate our approach on IAM~\cite{iam} and RIMES 2011~\cite{rimes}, two widely adopted benchmarks for both HTR and writer adaptation research~\cite{pippi2023choose,metahtr,docttt,wang2022fast,wordstylist}. IAM contains 13k English handwritten text lines from 657 writers, drawn from the Lancaster-Oslo/Bergen Corpus, while RIMES contains 12k French handwritten text lines from over 1,300 writers, consisting of administrative correspondence. Both datasets provide line-level samples with writer identity annotations, a requirement for our context-driven training protocol. For all experiments, we employ a CNN encoder composed of 18 convolutional layers followed by 12 depthwise separable convolutional layers, and a Transformer decoder with 8 layers, an embedding dimension of 256, and 4 attention heads. The overall architecture comprises approximately 8 million parameters. We train the model using the AdamW optimizer with a learning rate of $1\times10^{-4}$. To enhance robustness and generalization, we adopt curriculum learning, noise injection, and data augmentation strategies. At both training and inference time, \(K\) context lines are randomly sampled from the target writer within the same data split as the query image, without any pre-selection or filtering, ensuring a realistic and unbiased evaluation of the writer adaptation capability.

First, we introduce the results of the Context-Driven strategy applied to our CNN-Transformer model. 
Second, we compare the two training methods, Context-Driven and standard OCR (without context).
Finally, we investigate the fusion of the predictions from two models trained using each strategy and compare the resulting performance with the state-of-the-art results on the IAM and RIMES datasets. 

\subsection{Context-Driven evaluation}
\vspace{-1mm}

We first evaluate our context-driven model’s capacity to fulfill its training objective: predicting characters conditioned on their context encoding and generating \(\langle \text{ooc}\rangle\)  tokens whenever a character is absent from the context. Performance is measured using the Token Error Rate (TER), calculated as the edit distance between predicted and ground-truth token sequences, normalized by the total number of tokens.  While computed similarly to the Character Error Rate (CER), TER reflects the additional challenge of correctly recognizing when contextual information is missing through the prediction of \(\langle \text{ooc}\rangle\)tokens.

\begin{figure}[h]
  \centering
  \resizebox{10cm}{!}
{\includegraphics{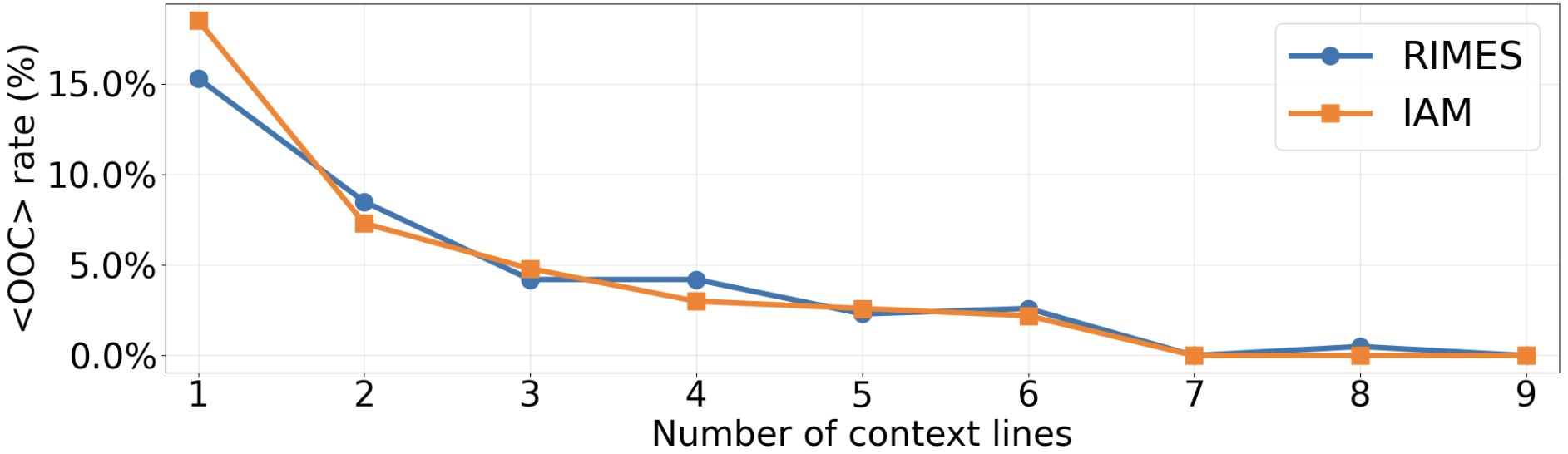}}
\vspace{-5mm}

\caption{ \(\langle \text{out-of-context}\rangle\) token rate vs. number of context lines for IAM and RIMES.}
  \label{fig:ooc_rate}
\end{figure}

To effectively apply this context-driven strategy, the context must provide sufficient coverage of the characters present in the query; otherwise, the model would predominantly predict \(\langle \text{ooc}\rangle\)  tokens. Let \(K\) be the number of context lines. Due to the natural redundancy of characters in written English and French, even a small \(K\) provides substantial coverage of the query characters. On the IAM and RIMES datasets, using only two context lines reduces the prediction rate of \(\langle \text{ooc}\rangle\)  tokens to less than 10\% (see Fig.~\ref{fig:ooc_rate}).  Increasing the context to seven lines makes  \(\langle \text{ooc}\rangle\) predictions almost entirely negligible. 

\vspace{-5mm}

\vspace{-10mm}

\begin{table}[h]
\caption{Analysis of the Context-Driven model as a function of $K$, the
maximum number of context lines, on IAM and RIMES.
\textbf{TER}: Token Error Rate;
\textbf{F1 $\langle$ooc$\rangle$}: F1-score of the $\langle$ooc$\rangle$ token prediction.}
\vspace{-1mm}
\centering
\resizebox{7.5cm}{!}{%
\begin{tabular}{c|cc|cc}
\toprule
 & \multicolumn{2}{c|}{\textbf{IAM}} & \multicolumn{2}{c}{\textbf{RIMES}} \\
\midrule
\textbf{K}
  & \textbf{TER (\%)} & \textbf{F1 $\langle$ooc$\rangle$ (\%)} &  \textbf{TER (\%)} & \textbf{F1 $\langle$ooc$\rangle$ (\%)}  \\
\midrule
1 & 6.30 & 92.21  & 5.13 & 92.60  \\
3 & 4.99 & 88.20  & 3.15 & 93.14  \\
5 & 4.73 & 85.34  & 2.91 & 91.18  \\
7 & 4.69 & 83.26  & 2.87 & 89.92  \\
9 & 4.64 & 82.61 & 2.92 & 88.69  \\
\bottomrule
\end{tabular}%
}
\label{tab:context_driven_per_line}
\end{table}
\vspace{-3mm}
Table~\ref{tab:context_driven_per_line} shows the impact of the maximum number of context lines $K$ on IAM and RIMES. Increasing $K$ consistently reduces the Token Error Rate (TER), confirming the benefit of larger contextual windows. On IAM, TER decreases from 6.30\% ($K=1$) to 4.64\% ($K=9$), while on RIMES it drops from 5.13\% to 2.87\%, with the best result at $K=7$, highlighting improved character disambiguation through in-context examples. The F1-score of the $\langle\text{ooc}\rangle$ token decreases as $K$ grows (from 92.21\% to 82.61\% on IAM and from 92.60\% to 88.69\% on RIMES). As the context expands, distinguishing truly out-of-context characters becomes more challenging. However, as seen in Fig. \ref{fig:ooc_rate}, increasing $K$ substantially reduces the number of $\langle\text{ooc}\rangle$ occurrences, making the F1-score more sensitive to small variations in prediction errors. 

Regarding transcription noise, real-world conditions such as struck-through characters, transcription errors, or imperfect line segmentation can degrade the $X_c$/$Y_c$ correspondence. We empirically observe that recognition performance remains largely stable under such moderate perturbations, which we attribute to the natural presence of noisy samples in the training data 
combined with our noise injection strategy introduced in section \ref{sec:train_protocol}.

\subsection{Standard OCR versus Context-Driven}
\label{sec:complementary}
In this section, we compare the Context-Driven and standard OCR training strategies using the same CNN-Transformer backbone and an identical parameter count, ensuring a fair comparison as described in Section~\ref{sec:setup}.  Each model is trained on the target dataset, either IAM or RIMES. The standard OCR models achieve state-of-the-art performance with Character Error Rates (CER) of 2.94\% on RIMES and 4.22\% on IAM.

\begin{figure}[h]
  \centering
  \resizebox{6.5cm}{!}{\includegraphics{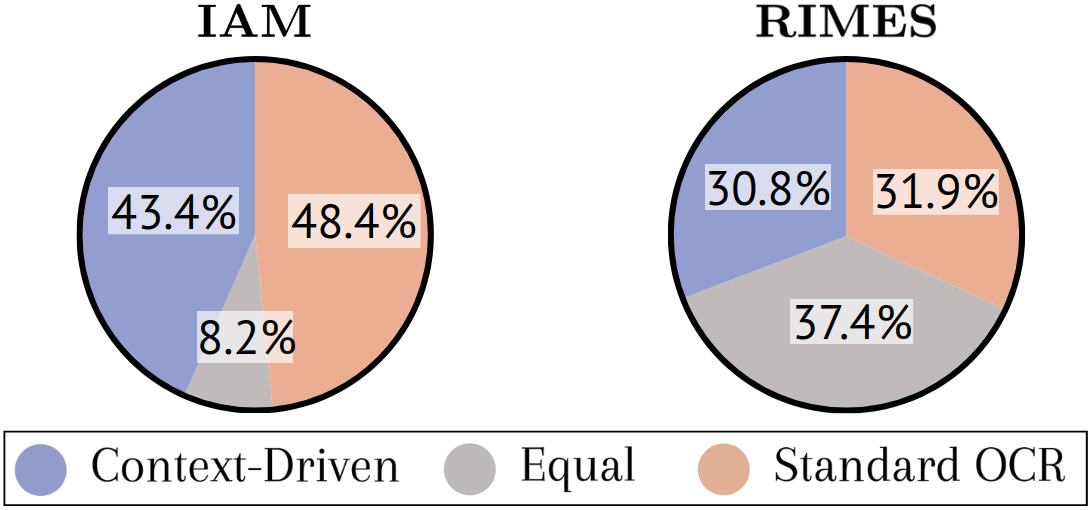}}
\caption{Distribution of writers according to which model achieves the lowest Character Error Rate (CER) on fully context-covered lines.  ``Equal'' denotes identical CER for both models. Maximum of 9 context lines.}  \label{fig:writer_percent}
\end{figure}

At the writer level, when evaluating CER fully context-covered lines (i;e without $\langle\text{ooc}\rangle$ tokens), results indicate that the effectiveness of the models is writer-dependent (see Fig.~\ref{fig:writer_percent}). Some writers achieve lower CER with the Context-Driven model, while others perform equally well or better with the Standard OCR model. This suggests that the two approaches are complementary: in-context examples improve recognition for certain handwriting styles, whereas baseline OCR is better for others.

\vspace{-5mm}
\vspace{-5mm}

\begin{figure}[h]
  \centering
  \resizebox{\textwidth}{!}{\includegraphics{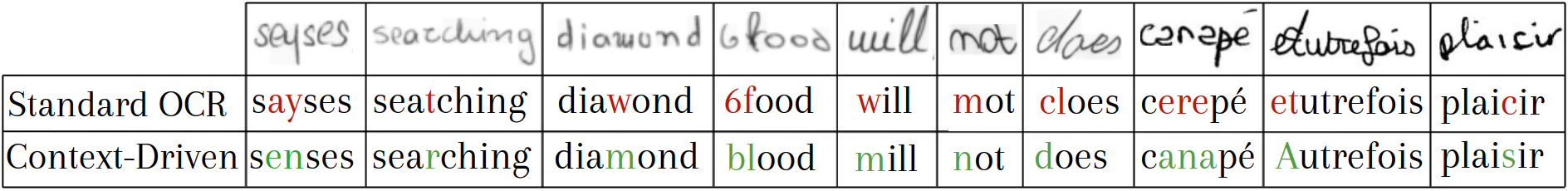}}
  \caption{Qualitative samples showing improvements obtained with context-driven few-shot adaptation over a baseline OCR model.}
  \label{fig:icl_better_ocr}
\end{figure}

We observe that, for ambiguous handwritten characters, the Context-Driven model leverages in-context examples to correct errors that the standard OCR model fails to resolve, demonstrating the complementary strengths of contextual adaptation (see Fig.~\ref{fig:icl_better_ocr}). These results indicate that even for a state-of-the-art OCR system, leveraging in-context examples can provide substantial improvements for a significant proportion of writers.

\vspace{-5mm}
\vspace{-10mm}

\subsection{Fusion of Standard OCR and Context-Driven Training}
Motivated by the complementary strengths of the two methods, we fuse their predictions as described in Section~\ref{sec:fusion}.  

\vspace{-3mm}
\subsubsection{Impact of Context Length}
To evaluate the scalability of our context-driven approach, we investigate the influence of the number of context lines on the fusion performance (see Table~\ref{tab:fusion_per_line}).

\vspace{-4mm}
\begin{table}[h]
\small
\caption{Effect of the number of context lines $K$ on fusion performance 
on IAM and RIMES. The row at $K=0$ corresponds to the standard OCR 
model used as baseline, without any fusion.}
\centering
\small
\begin{tabular}{c|cc|cc}
\toprule
 & \multicolumn{2}{c|}{\textbf{IAM}} & \multicolumn{2}{c}{\textbf{RIMES}} \\
\textbf{K} & CER (\%) & WER (\%) & CER (\%) & WER (\%) \\
\midrule
0 & 4.22 & 14.10 & 2.91& 8.67 \\
\midrule
1 & 4.20 & 14.60 & 2.83& 9.91\\
3 & 4.00 & 13.96 & 2.47& 8.86\\
5 & 3.96 & 14.03 & 2.41& 8.38\\
7 & 3.94 & 13.93 & 2.35& 8.08\\
9 & \textbf{3.92} & \textbf{13.81} & \textbf{2.34} & \textbf{8.01}\\
\bottomrule
\end{tabular}
\vspace{1mm}

\label{tab:fusion_per_line}
\end{table}
\vspace{-10mm}

\begin{itemize}
    \item \textbf{Consistent Error Reduction:} We observe a clear downward trend in error rates as the number of context lines increases. On the IAM dataset, increasing $K$ from 1 to 9 leads to a steady reduction in CER from 4.20\% to 3.92\%. A similar and even more pronounced trend is visible on RIMES, where the CER drops from 2.83\% ($K=1$) to 2.34\% ($K=9$), representing a 19.5\% relative improvement over the baseline of the standard OCR ($K=0$).
    
    \item \textbf{CER vs. WER Dynamics:} On the RIMES dataset, we observe that at $K=1$, while the CER improves slightly compared to the baseline, the WER initially increases (from 8.67\% to 9.91\%). This suggests that with very limited context, the model may over-correct character-level predictions in a way that disrupts word-level coherence. However, as $K$ increases ($K \geq 5$), the WER drops significantly below the baseline, reaching 8.01\% at $K=9$. This demonstrates that a larger \textbf{context length} is necessary for the model to improve character predictions while preserving word-level integrity.
\end{itemize}

To validate our confidence-based selection strategy, we compare it against a random fusion baseline that randomly selects at each aligned position either the Context-Driven or the Standard OCR prediction, while preserving the special $\langle\text{ooc}\rangle$  substitution rule. Confidence-based fusion consistently outperforms random fusion, achieving a CER of 3.92\% versus 4.83\% on IAM and 2.34\% versus 2.82\% on RIMES with \(K=9\) context lines, confirming that the confidence scores provide a meaningful signal for selecting the most reliable prediction at each character position. We note that the fusion strategy could be further enhanced by incorporating language-aware refinement mechanisms or integrating an external language model to improve linguistic coherence. Nevertheless, the primary objective of this work is to demonstrate the benefits of stronger visual grounding through multimodal in-context examples for writer adaptation. 
Overall, the results demonstrate that our fusion approach effectively combines the complementary strengths of context-driven adaptation and standard OCR decoding, ultimately surpassing the performance of standalone HTR models that do not perform writer adaptation.

\vspace{-3mm}
\subsubsection{Comparison with State-of-the-Art}
The performance of the proposed fusion strategy is evaluated against state-of-the-art (SOTA) methods on the IAM and RIMES datasets, as summarized in Table~\ref{tab:sota}. On IAM and RIMES, our method achieves Character Error Rates (CER) of 3.92\% and 2.34\%, respectively, significantly outperforming standard OCR models without test-time adaptation (TTA). For a fair comparison, we included only SOTA models with up to 100 million parameters—with Kang et al. (100M) and TrOCR-small (62M) being the largest, while all other baselines contain fewer than 20 million parameters.

\vspace{-5mm}
\begin{table}[h]
\small
\centering
\caption{Line-level comparison of handwritten text recognition (HTR) methods on the IAM and RIMES datasets. ``--'' indicates values not reported in the original publication; $^{*}$ denotes paragraph-level evaluation. \textit{Ours-baseline} refers to our model without any writer-specific adaptation. MAE denotes a Masked Autoencoder-based image reconstruction auxiliary decoder.}
\vspace{-1mm}
\begin{tabular}{l@{\hspace{0.25cm}}lccccc}
\toprule
\multirow{2}{*}{\textbf{Method}} 
& \multirow{2}{*}{\textbf{Architecture}} 
& \multicolumn{2}{c}{\textbf{IAM}} 
& \multicolumn{2}{c}{\textbf{RIMES}} 
& \multirow{2}{*}{\textbf{\shortstack{Param.\\updates}}} \\
& & \textbf{CER} & \textbf{WER} & \textbf{CER} & \textbf{WER} & \\
\midrule
\multicolumn{7}{l}{\textit{Without test-time adaptation}}\\
SPAN~\cite{span} 
& FCN          & 5.45\% & 19.83\% & 3.81\% & 13.8\%  & No\\
Decoupled AN~\cite{decoupledan} 
& CNN-RNN + Attn     & 6.4\%  & 19.6\%  & 2.7\%  & 8.9\%   & No\\
Kang et al.~\cite{kang2022pay} 
& CNN-Transf.    & 4.67\% & 15.45\% & --   & --    & No\\
Retsinas et al.~\cite{bestpractices} 
& CNN-BiLSTM   & 4.62\% & 15.89\% & 2.75\% & 9.93\%  & No\\
VAN~\cite{van} 
& CNN-LSTM + Attn    & 4.32\% & 16.24\% & 3.04\% & 8.32\%  & No\\
VLT~\cite{vlt} 
& CNN-Transf.    & 4.23\%   & 14.61\%    & 2.82\% & 8.47\%  & No\\
DAN~\cite{dan} 
& CNN-Transf.    & --   & --    & 2.63\% & 6.78\%  & No\\
TrOCR-small~\cite{trocr} 
& ViT- Transf.& 4.22\% & --    & --   & --    &  No\\
 \textbf{Ours (baseline)}& CNN-Transf.& 4.22\%& 14.10\%& 2.91\%& 8.67\%& No\\

\midrule
\multicolumn{7}{l}{\textit{With Test-time adaptation}} \\
DocTTT \cite{docttt} 
& CNN-Transf. + MAE    
& 4.22\%$^{*}$ & 14.17\%$^{*}$ & 2.33\% & 6.47\%  & Yes \\
MetaWriter~\cite{metawriter} 
& CNN-Transf. + MAE   & 3.32\% & 10.21\% & 2.13\% & 6.55\%  & Yes \\
Wang et al.~\cite{wang2022fast} 
& CNN-RNN + Attn     & 5.3\%  & 18.5\%  & --   & --    & \textbf{No}  \\
\textbf{Ours} 
& CNN-Transf.    
& {3.92\%} & {13.81\%} 
& {2.34\%}& {8.01\%}& \textbf{No}  \\
\bottomrule
\end{tabular}
\label{tab:sota}
\end{table}
\vspace{-7mm}

When evaluated against adaptive methods that strictly operate without parameter updates, our approach surpasses Wang et al.~\cite{wang2022fast}, demonstrating that explicit visual grounding through in-context examples is a more effective strategy than global style embedding extraction for parameter-free writer adaptation. Overall, our method establishes a new 
state-of-the-art among parameter-free writer adaptation approaches, surpassing all HTR models that do not perform writer-specific adaptation at inference time.

Regarding DocTTT~\cite{docttt} and MetaWriter~\cite{metawriter} both perform gradient-based parameter updates at inference time through an auxiliary reconstruction objective. Our approach attains comparable performance to DocTTT on RIMES (2.34\% vs. 2.33\%) and outperforms it on IAM at the line level, without any parameter updates. Regarding MetaWriter, as discussed in Section~\ref{sec:writer_adaptation}, coordinating reconstruction and recognition objectives requires careful tuning of multiple interdependent hyperparameters. Despite extensive tuning of the adaptation learning rate, masking strategy, number of update steps, and reconstruction loss weighting, we could not fully match the reported results under our setup and thus refer to the original publication for comparison. 

Overall, our approach demonstrates that strong writer adaptation performance can be achieved through a simple, parameter-free mechanism based on multimodal in-context examples, setting a new state-of-the-art among methods without test-time updates while remaining competitive with more complex MAE-based adaptation strategies.

\vspace{-2mm}
\section{Conclusion}

In this work, we proposed a novel context-driven writer adaptation framework inspired by Multimodal In-Context Learning (MICL). In contrast to offline fine-tuning — which requires retraining before deployment — and MAE-based methods that perform gradient updates at inference time, our approach conditions predictions on a small support set of image-text pairs $(X_c, Y_c)$ from the target writer, replacing implicit reconstruction objectives with an adaptation mechanism directly grounded in character-level visual pattern matching, without any parameter updates. Building upon the principles introduced in Rosetta, we mitigate the well-known modality imbalance in MICL by strictly conditioning predictions on multimodal in-context examples and demonstrate that a compact 8-million-parameter CNN-Transformer is sufficient to achieve strong in-context adaptation capabilities. Our empirical analysis shows consistent performance gains as context length increases from 1 to 9 lines, and that combining context-driven and standard OCR predictions yields complementary improvements, achieving state-of-the-art Character Error Rates of 3.92\% on IAM and 2.34\% on RIMES.

A current limitation of our approach is the requirement for labeled context transcriptions $Y_c$ at inference time. However, we argue that this constraint remains practically lightweight: fewer than ten annotated lines from the target writer are sufficient to achieve meaningful adaptation. More broadly, the idea of personalizing an HTR model to a new writer using only a handful of annotated samples — without any retraining or parameter updates — represents an attractive and scalable paradigm for real-world deployment, where collecting a small number of labeled examples is often feasible and far less demanding than traditional fine-tuning pipelines.
Overall, this work establishes that multimodal in-context learning can be effectively leveraged as a lightweight and scalable framework for few-shot writer adaptation at inference time, without requiring any parameter updates.

\section{Acknowledgment}
The present work was performed using computing resources of CRIANN (Regional HPC Center, Normandy, France).  This work was financially supported by the French Defense Innovation Agency and by the Normandy
region. 
\vspace{-2mm}
\begin{figure}[h]
  \centering
  \resizebox{6.5cm}{!}{\includegraphics{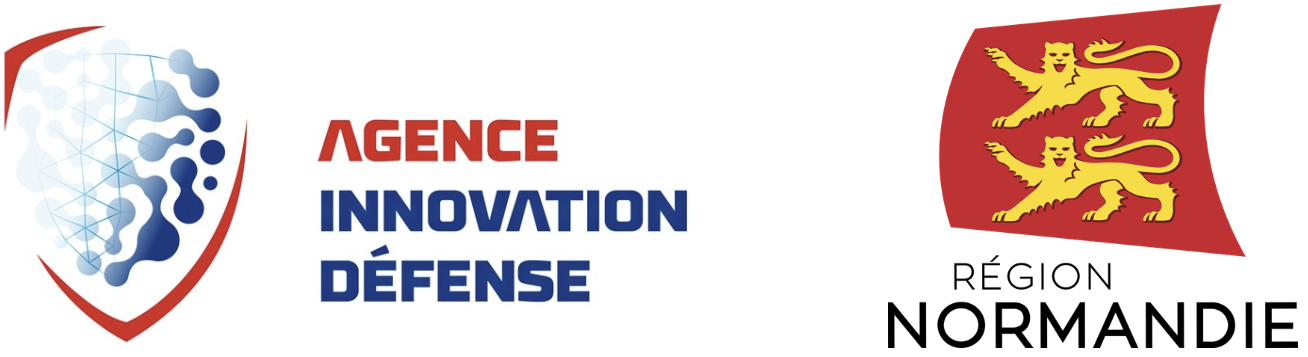}}
\label{fig:fusion}
\end{figure}
%
%
%
%

\bibliographystyle{splncs04}  
\bibliography{references}

\end{document}